\documentclass[runningheads]{llncs}

\usepackage{eccv}

\usepackage{eccvabbrv}

\usepackage{graphicx}
\usepackage{booktabs}
\usepackage{enumitem}
\usepackage{wrapfig}
\usepackage{multirow}
\usepackage{makecell}

\usepackage[accsupp]{axessibility}  

\usepackage[pagebackref,breaklinks,colorlinks,citecolor=eccvblue]{hyperref}

\usepackage{orcidlink}

\begin{document}

\title{My3DGen: A Scalable Personalized 3D Generative Model} 

\titlerunning{My3DGen}


\author{Luchao Qi$^{1}$\hspace{0.5em}\and
Jiaye Wu$^{2}$\and
Annie N. Wang$^{1}$\and
Shengze Wang$^{1}$\and
Roni Sengupta$^{1}$}

\authorrunning{Qi et al.}

\institute{University of North Carolina at Chapel Hill\\
\email{\{lqi, awang13, shengzew, ronisen\}@cs.unc.edu}\\
\and
University of Maryland, College Park\\
\email{jiayewu@cs.umd.edu}
}

\maketitle

\begin{abstract}
In recent years, generative 3D face models (e.g., EG3D) have been developed to tackle the problem of synthesizing photo-realistic faces.
However, these models are often unable to capture facial features unique to each individual, highlighting the importance of personalization.
Some prior works have shown promise in personalizing generative face models, but these studies primarily focus on 2D settings.
Also, these methods require both fine-tuning and storing a large number of parameters for each user, posing a hindrance to scalability.
Another challenge of personalization is the limited number of training images available for each individual, which often leads to overfitting when using full fine-tuning methods.
Our proposed approach, My3DGen, generates a personalized 3D prior of an individual using as few as 50 training images.
My3DGen allows for novel view synthesis, semantic editing of a given face (e.g. adding a smile), and synthesizing novel appearances, all while preserving the original person's identity.
We decouple the 3D facial features into global features and personalized features by freezing the pre-trained EG3D and training additional personalized weights through low-rank decomposition.
As a result, My3DGen introduces only \textbf{240K} personalized parameters per individual, leading to a \textbf{127}$\times$ reduction in trainable parameters compared to the \textbf{30.6M} required for fine-tuning the entire parameter space.
Despite this significant reduction in storage, our model preserves identity features without compromising the quality of downstream applications.
\keywords{Personalization \and 3D-GAN \and 3D Face}
\end{abstract}
    
\vspace{-3em}
\section{Introduction}
\vspace{-1em}
\label{sec:intro}
Recently, dramatic advancements in deep generative models like generative adversarial networks (GANs)~\cite{pmlr-v70-arjovsky17a, park2019SPADE, karras2018progressive, radford_unsupervised_2016} and diffusion models~\cite{ho_denoising_2020, dhariwal_diffusion_2021} have led to a surge in their popularity for computer vision applications. Notably, GANs have proven to be particularly powerful at generating realistic photos of faces~\cite{karras_progressive_2018, karras2019style, karras2020analyzing, karras2020training, karras2021alias}. These techniques have been extended to 3D vision as well, leading to the development of models that are capable of reconstructing 3D models of existing 2D facial images, synthesizing novel appearances, and editing various facial attributes such as facial expressions ~\cite{niemeyer_giraffe_2021, schwarz_graf_2020, nguyen2019hologan, chan_efficient_2022, or2022stylesdf}.
These models play a crucial role in enhancing the authenticity of virtual communication, AR/VR/MR, and content creation, thereby increasing engagement. 

\vspace{-2em}
\begin{figure}[th]
  \centering
  \includegraphics[width=\linewidth]{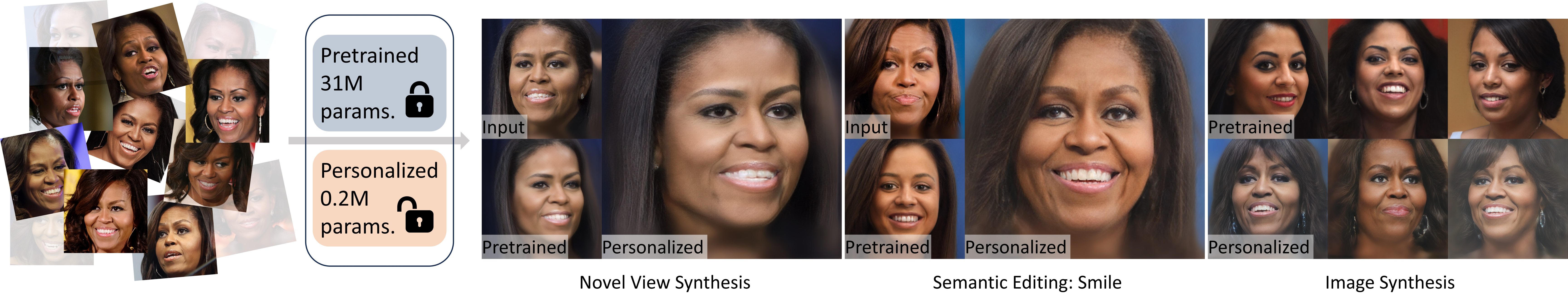}
    \vspace{-2em}
    \captionof{figure}{
    Given 50 images of \textit{Michelle Obama}, we personalize a pre-trained 3D generative prior and demonstrate the applications in various downstream tasks.
    Each downstream task presents the original input image of \textit{Michelle} (top left), alongside the corresponding output generated using the pre-trained face prior (bottom left), compared to the output using our personalized face prior (right). 
    Our personalized prior can faithfully retain the key facial characteristics of \textit{Michelle Obama}, as opposed to the pre-trained prior.
    \vspace{-2em}
    }
  \label{fig:teaser}
\end{figure}

However, current 3D generative models~\cite{niemeyer_giraffe_2021, schwarz_graf_2020, nguyen2019hologan, chan_efficient_2022, or2022stylesdf} are unable to create authentic 3D face models of particular subjects.
Although these models can synthesize photo-realistic fake faces, they cannot generate, reconstruct, or modify the distinctive traits of a particular real person's face without distorting their identity. This problem is further exaggerated for underrepresented demographics whose facial characteristics are sparsely represented in widely used training datasets like FFHQ~\cite{karras2019style} or CelebA~\cite{liu2015deep}.
Existing inversion techniques~\cite{bhattarai2024triplanenet, trevithick2023real, ko20233d, xie2023high, roich2022pivotal, abdal20233davatargan} are only able to preserve identity by tuning the model separately for every test image, an impractical and inefficient approach. Furthermore, they cannot edit the inverted image or synthesize novel appearances without identity distortion.
Thus, this paper presents an approach to create a personalized 3D generative prior for an individual. This prior enables 3D facial reconstruction, synthesis of novel appearances, and editing of existing appearances, while maintaining the individual's identity.

A major obstacle to the deployment of personalized 3D generative models at scale for real-world applications is their enormous storage demand. We can illustrate this scalability problem with a concrete example: Consider that naively personalizing a 3D generative model such as EG3D~\cite{chan_efficient_2022} requires storing approximately 31 million parameters (121 MB) for each user. For three billion users (monthly active users of Facebook), this would require 363 PB of memory, an extremely cost-prohibitive demand even for a company the size of Meta.
Evidently, we need to design more parameter-efficient personalization techniques that will enable the building of 3D generative priors for a large population. Another challenge of personalization is that an average user often takes only a limited number of photos of themselves each day or week~\cite{miller_give_2007}. The limited training set size makes it difficult to fine-tune a large global generative model, often leading to overfitting, mode collapse~\cite{aghabozorgi_adaptive_2023}, and data drift~\cite{lee_countering_2019, lu_countering_2020}, all of which prevent the model from generalizing to unseen test images of an individual.


Our idea is to decouple the facial features of an individual into (a) shared global features that can be represented by a generative model trained across many different identities and (b) personalized features of an individual that can be represented with much fewer trainable parameters, trained on images of that individual only. We use EG3D~\cite{chan_efficient_2022} as our pre-trained generative model and train additional weights for low-rank decompositions of every convolutional and fully-connected layer to capture the personalized features of that individual. Drawing inspiration from the recent success of Low-Rank Adaption (LoRA)~\cite{hu_lora_2021} in parameter-efficient fine-tuning of large language models~\cite{hu_llm-adapters_2023,yu_differentially_2022} and diffusion models~\cite{ruiz_hyperdreambooth_2023}, we use LoRA during personalization, which hasn't been well explored in convolution-heavy GAN-based models before. Our approach allows for personalization using only 240K (0.9 MB) trainable parameters instead of requiring fine-tuning of all 31 million (121 MB) parameters. For three billion users, this means that only 2.7 PB of storage memory would be needed instead of 363 PB. 

Quantitative and qualitative analyses show that our proposed My3DGen outperforms the pre-trained 3D generative model EG3D~\cite{chan_efficient_2022} on multiple tasks including 3D reconstruction, novel appearance synthesis, image enhancement, and semantic editing. Additionally, our personalized model can produce results similar to those achieved by naively fine-tuning a pre-trained model with 31 million parameters~\cite{chan_efficient_2022}, while only using as few as 240K trainable parameters. 
We further provide insights based on different parametrization strategies. We observe that increasing the rank of LoRA modules only contributes to better overfitting of the background or small stylistic changes without improving the shape or the identity-preserving performance after personalization. Furthermore, we find that personalizing the earlier layers of StyleGAN2 has the most impact on the resulting quality, indicating that coarse and middle layers are more responsible for capturing the shape and identity of the person. 

In conclusion, we propose the following contributions. (1) To the best of our knowledge, our research presents the first attempt to personalize a 3D generative model and demonstrates its effectiveness in various downstream tasks;
(2) We design a generative prior where an individual's facial features are disentangled into global features--represented by a pre-trained 3D generative model, and personalized features--captured by training additional low-rank weights requiring only 240K stored parameters. This method helps to avoid overfitting and has the potential to improve performance over naive tuning without LoRA; (3) Departing from previous works that primarily utilized LoRA solely for linear layers within transformers or diffusion models, our approach incorporates LoRA into convolution-centric GAN architectures, presenting an innovative perspective on using LoRA for 3D personalization.

\vspace{-1.5em}
\section{Related Work}
\vspace{-1em}
\label{sec:related_work}
\noindent
\textbf{3D Face Reconstruction.}
Parametric 3D models, also known as 3D Morphable Models (3DMM)~\cite{3dmm, animatable_3DMM}, are often widely used for modeling 3D human faces through a linear combination of basic shapes. Typically, 3DMMs are constructed using high-quality facial scans from multiple individuals. However, when fitting a 3DMM to a test image~\cite{lattas_avatarme_2022, tran_regressing_2017, tran2019learning}, the result often appears unrealistic due to limitations in the expressiveness of the linear blended model. Also, training a 3DMM using 2D photo collections is challenging due to the absence of corresponding 3D scans. As a result, recent research has focused on developing generative models for 3D faces~\cite{chan_pi-gan_2021, gu_stylenerf_2021, or2022stylesdf}, which are able to achieve a remarkable level of facial detail and capture small wrinkles and bumps by training solely on 2D images. Nevertheless, these models only aim to generate arbitrary fake faces, and there is a lack of research addressing the personalization of such models, which is the focus of our paper.

\noindent
\textbf{Personalized Generative Models.}
Personalization plays a significant role in the field of generative AI, with diverse applications including deepfakes, video avatars, talking heads, and text-to-image (T2I) generation~\cite{westerlund2019emergence,yao_dfa-nerf_2022,ruiz_hyperdreambooth_2023, choi_personalized_2023,ruiz2023dreambooth, paruchuri_motion_2023, cheng2022videoretalking, aghabozorgi_adaptive_2023}. 
Previous inversion works~\cite{alaluf2021restyle, bhattarai2024triplanenet, roich2022pivotal} designed a model-dependent encoder or fine-tuned the pre-trained model to best fit an input image. 
However, these methods do not possess the ability to generate diverse appearances or edit existing appearances of an individual without distorting their identity. Most inversion methods also require updating and storing a separate set of network weights for every single image, which is impractical. In personalizing a 3D generative model, our goal is to use only a single set of network weights that can reconstruct, edit, and synthesize any novel appearances of an individual while preserving identity. Our work is inspired by the recent success in 2D generative model personalization~\cite{nitzan2022mystyle, mystyle++}. However, 2D personalization cannot create 3D faces or produce unique perspectives, and frequently falters for non-frontal positions due to a lack of geometric information. Our paper extends personalization from 2D to 3D by assuming that an individual's facial appearance can be decomposed into global features represented by the pre-trained model and personalized characteristic features represented by parameter-efficient low-rank adaptive weights.

\noindent
\textbf{Parameter-Efficient Fine-Tuning.} Large foundation models~\cite{radford2019language, roberta, gpt3, rombach2022high, dosovitskiy_image_2021} often achieve impressive performance for tasks in their domain. However, the huge number of parameters in such models often prevents them from being fine-tuned for downstream tasks using a limited budget. For example, GPT-4~\cite{openai_gpt-4_2023} contains 1.76 trillion parameters, which is impractical for most users to fine-tune. To tackle this issue, many parameter-efficient fine-tuning (PEFT) techniques have been previously proposed to fine-tune models efficiently. In natural language processing, approaches~\cite{houlsby2019parameter, hu_lora_2021, lin-etal-2020-exploring} have been proposed to enable efficient adaptation of pre-trained language models to various downstream applications without fine-tuning all of the model's parameters. For image generation, ControlNet~\cite{Zhang_2023_ICCV}, HyperDreamBooth~\cite{ruiz_hyperdreambooth_2023}, and AnimateDiff~\cite{guo_animatediff_2023} have been proposed to tune pretrained diffusion models with additional modules.
Our approach is inspired by LoRA~\cite{hu_lora_2021}, a technique for efficiently finetuning large language foundation models by imposing low-rank structures on weight matrices. LoRA~\cite{hu_lora_2021} was first proposed as a way to predict additional network weights without changing the pre-trained transformer, allowing efficient adaptation and storage for task-specific models~\cite{hu_lora_2021}. Subsequently, LoRA has found widespread use in fine-tuning pre-trained networks for various downstream tasks, including network quantization, parameter budget allocation, and continual learning~\cite{dettmers_qlora_2023,zhang_adaptive_2023,smith_continual_2023}. 
\vspace{-2em}
\section{Method}
\begin{wrapfigure}{r}{0.5\linewidth}
\vspace{-3em}
    \centering
    \includegraphics[width=\linewidth]{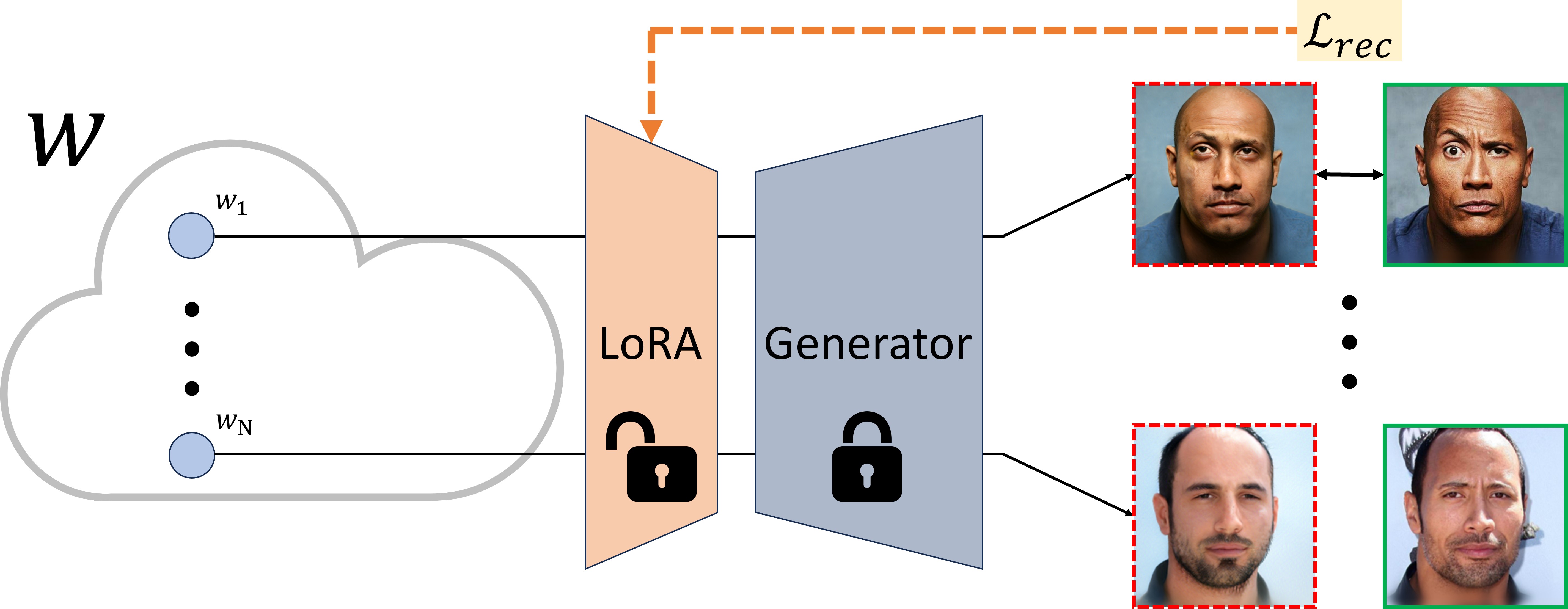}
    \vspace{-1em}
    \captionof{figure}{Architecture of our personalization approach. We project an individual's images into StyleGAN2's $\mathcal{W}$ space through latent code optimization to obtain a set of latent anchors. We then tune the generator to reconstruct an individual's images. During tuning, the generator is frozen while only LoRA weights are personalized.}
    \label{fig:architecture}
    \vspace{-2em}
\end{wrapfigure}

\vspace{-1.5em}
Our objective is to personalize a 3D generative model with ideally 50 images of an individual.
First, we introduce background concepts in Sec.~\ref{method:preliminaries}. Then we formulate the problem of personalizing EG3D~\cite{chan_efficient_2022}, a pre-trained 3D generative model, in Sec.~\ref{method:personalization}. Finally, we discuss how we incorporate parameter-efficient personalization into our model in Section~\ref{method:param_efficient} with a visual overview in Fig.~\ref{fig:architecture}. 

\vspace{-1.5em}
\subsection{Preliminaries}
\vspace{-0.5em}
\label{method:preliminaries}
\noindent
\textbf{EG3D.}
In this work we use the state-of-the-art 3D generative model EG3D~\cite{chan_efficient_2022} as the pre-trained model that captures global facial features across multiple identities.
EG3D has four main components: (i) a StyleGAN2 generator that takes as input a random latent code and outputs $256\times256\times96$ feature maps, which are further reshaped into three $256\times256\times32$ triplanes, (ii) a neural renderer that decodes triplane features and renders a face given a camera pose input, (iii) a super-resolution module that upsamples rendered images to a resolution of $512\times512$, and (iv) a StyleGAN2 discriminator that differentiates between generated images and real images. 
The entire pipeline is trained following the typical non-saturating minimax GAN loss~\cite{goodfellow2020generative}.
In this paper, we discard the discriminator and will use a reconstruction loss to personalize the generator.

\noindent
\textbf{LoRA.}
Our approach is based on Low-Rank Adaptation of Large Language Models (LoRA)~\cite{hu_lora_2021}, a technique for efficiently finetuning convolution-free transformer networks by imposing low-rank structures on weight matrices.
LoRA shows that a pre-trained model's weight matrix $ W_{0} \in \mathbb{R}^{d \times {k}} $ for any fully-connected layer can be fine-tuned with a low-rank decomposition using the following formulation: Let $ W_\text{ft} = W_{0}+\Delta W=W_{0}+B A $ be the fine-tuned adapted weight matrix, where $ B \in \mathbb{R}^{d \times r}$ and $ A \in \mathbb{R}^{r \times k} $ are trained while $W_{0}$ is frozen, and the rank $r \ll \min (d, k)$~\cite{hu_lora_2021}.
Although this approach demonstrates impressive decomposition performance for transformers~\cite{he2023parameter} and diffusion models~\cite{ruiz_hyperdreambooth_2023}, previous work solely focuses on the decomposition of linear layers~\cite{wang_customizing_2023,guo_animatediff_2023,ruiz_hyperdreambooth_2023,smith_continual_2023, kumari2023multi}, while GAN-based models are convolution-heavy. In our paper, we apply LoRA to the convolution-heavy EG3D model for personalization.


\vspace{-1.5em}
\subsection{Personalization Formulation}
\vspace{-0.5em}
\label{method:personalization}
We consider a scenario where an individual has a training set of $N$ 2D images \(x_i\), each with an associated camera pose \(c_i\), denoted as $\mathcal{D}_{p}=\left\{(x_{i}, c_{i})\right\}_{i=1}^{N}$. We let $G(\cdot;\theta_{G} + \Delta \theta_G)$ denote our personalized EG3D model where $G$ is the EG3D architecture, $\theta_G$ represents the frozen pretrained weights, and $\Delta \theta_G$ represents our trainable parameters. Our goal is to tune this model on a low-dimensional manifold in the \(\mathbf{W}\) latent space, dubbed a personal convex hull~\cite{nitzan2022mystyle}, which is a subspace defined by all latent codes of the $N$ images.
Our personalization scheme is inspired by the success of PTI~\cite{roich2022pivotal} and MyStyle~\cite{nitzan2022mystyle} in tuning a pretrained 2D StyleGAN. We first invert images from $(x_i, c_i) \sim \mathcal{D}_p$ with associated camera poses into latent vectors \(w_i\), dubbed anchors in MyStyle~\cite{nitzan2022mystyle}, with an off-the-shelf inversion technique~\cite{karras2020analyzing}.
That is, we freeze both the camera pose and the model weights while only optimizing the randomly initialized latent vector.

We then tune the weights $\Delta \theta_{G}$ of the model $G$ using reconstruction objective $\mathcal{L}_{rec}$, which represents the difference between each image $x_i$ and its reconstruction.
Formally,
\setlength{\abovedisplayskip}{3pt}
\setlength{\belowdisplayskip}{3pt}
\begin{align}
    \mathcal{L}_{rec}^{(i)} &= \mathcal{L}_{lpips}\left(G\left(w_{i}, c_{i};\theta_{G} + \Delta \theta_{G}\right), x_{i}\right) \nonumber \\
    &+ \lambda_{\mathcal{L}_{2}}\left\|G\left(w_{i},c_{i};\theta_{G} + \Delta \theta_{G}\right)-x_{i}\right\|_{2}
\end{align}
and we have $\Delta \theta_{G_p} = \operatorname*{argmin}_{\Delta \theta_{G}} \sum_{i=1}^N\mathcal{L}_{r e c}^{(i)}$ as our optimized parameters for the personalized EG3D model $G_p$. 


\vspace{-1.5em}
\subsection{Parameter Efficient Personalization}
\vspace{-0.5em}
\label{method:param_efficient}

Naively fine-tuning the model can lead to forgetting of knowledge learned in pretraining and entanglement of pre-trained facial features with the personalized features, compromising both the model's interpretability and generalization.
Therefore, in order to decouple the global features, captured in the pre-trained model, from the personalized features, we freeze the weights of the pre-trained model and tune additional weights to capture an individual's distinct facial priors. We use the technique of Low-Rank Adaptation (LoRA)~\cite{hu_lora_2021} to train the additional personalized weights, using only 240K parameters per identity. 

The original LoRA paper focuses on convolution-free transformers where linear fully connected layers are decomposed into low-rank submatrices $A$ and $B$, respectively~\cite{hu_lora_2021}. However, EG3D's generator StyleGAN2 and super-resolution modules are convolution-heavy, and convolution operations are often implemented with general matrix multiplication using the well-known `im2col' algorithm~\cite{anderson_high-performance_2020}. Thus, when adapting for personalization, we decompose both the convolution and fully connected layers in the StyleGAN2 generator and super-resolution modules using LoRA~\footnote{
We go into detail about our implementation of the convolutional LoRA decomposition in the supplementary material.
}. Pretrained weights \(W_0\) of the StyleGAN2 generator and the super-resolution module are frozen while only $A$ and $B$ are trained. We tune all the parameters of the neural renderer in EG3D, as it is relatively small with 4K parameters. Under this setting, the number of parameters needed to be tuned is determined by the rank $r$. With rank $r = 1$, we can reduce the 30.6M trainable parameters of EG3D to only 240k, a reduction of 127$\times$.
\vspace{-1.5em}
\section{Experiments}
\vspace{-1em}
We first discuss the details of our evaluation framework including datasets, experiment pipeline, and evaluation metrics in Sec.~\ref{exp:experiment_details}. Then we present both qualitative and quantitative results of our approach, My3DGen, in Sec.~\ref{exp:downstream}. We further analyze the effect of LoRA for personalization in Sec.~\ref{exp:lora_weight_change}.
Finally, we present ablation studies in Sec.~\ref{exp:ablation}.
 
\vspace{-1.5em}
\subsection{Experiment Details}
\vspace{-0.5em}
\label{exp:experiment_details}


\noindent
\textbf{Dataset.}
We conduct experiments using facial images of celebrities, the same dataset used by Mystyle~\cite{nitzan2022mystyle}. Images are preprocessed following~\cite{karras_progressive_2018, chan_efficient_2022} to align and crop faces to $512\times512$. Finally, the faces are separated into reference sets and test sets for each celebrity. The number of images included in the reference set and the test set for each celebrity is presented in the supplementary. Unless otherwise noted, for each celebrity, we personalize a model using 50 images from the reference set as the training set.

We use an off-the-shelf pose extraction~\cite{deng2019accurate} model to both identify the face region and label the pose of the face, following the same pipeline in EG3D~\cite{chan_efficient_2022}. We project images into the model's \(\mathcal{W}\) latent space following the StyleGAN~\cite{karras2020analyzing} optimization scheme for 500 iterations to obtain their latent codes (anchors).

\noindent
\textbf{Training.}  We choose 50 images per individual for all personalization tunings. The effect of dataset size is further discussed in Sec.~\ref{exp:ablation}. We tune the model starting from the pretrained EG3D (on FFHQ), which outputs images at resolutions of both $128\times128 \ (\boldsymbol{I}_{128})$ and $512\times512 \ (\boldsymbol{I}_{512})$. Hyper-parameters are chosen as follows: $\lambda_{lpips} = \lambda_{\mathcal{L}_{2}} = 1$. We apply $\mathcal{L}_{rec}$ on both $\boldsymbol{I}_{128}$ and $\boldsymbol{I}_{512}$. Our further experiments show that a higher LPIPS regularization weight ($\lambda_{lpips} > 1$) will lead to checkerboard-style artifacts and a lower LPIPS weight ($\lambda_{lpips} < 0.1$) will cause nonphotorealistic artifacts, which aligns with results in previous work~\cite{bhattarai2024triplanenet}. Based on this observation, we use $\lambda_{lpips} = 1$.

\noindent
\textbf{Evaluations.}
We evaluate our methods for the following primary downstream tasks: 1) \textit{Image inversion} through PTI~\cite{roich2022pivotal}; 2) \textit{Image interpolation} where we interpolate between two anchors in the latent space and generate images that morph from one face to another; 3) \textit{Image synthesis} where the goal is to generate novel appearances of an individual by sampling from a latent space, following the protocol outlined in Mystyle~\cite{nitzan2022mystyle}; 4) \textit{Image enhancement} tasks such as image-inpainting and super-resolution; 5) \textit{Semantic editing} where the goal is to modify the facial expression or age of the person while maintaining the identity and pose. Note that we modify the model weights after personalization via PTI only for image-inversion tasks, for consistency with prior research on 3D GAN~\cite{chan_efficient_2022}.
For any novel views, we render faces with a yaw range of $\pm 0.35$ (radians) and a pitch range of $\pm 0.25$ (radians) relative to the frontal face for all our experiments. Unless otherwise specified, we evaluate the performance of the personalized model on unseen test images. As our experiments were conducted in a 3D environment, we strongly suggest the reader refer to the supplementary video materials.

\noindent
\textbf{Metrics.} 
\label{exp:metrics}
When evaluating inversion outcomes, it is standard to use pixel-based metrics such as PSNR and SSIM or neural network-based perceptual metrics such as LPIPS~\cite{zhang_unreasonable_2018} and DISTS~\cite{ding_image_2022} to assess the inverted image and compare it with the original image. However, due to errors in the estimation of face poses, Live3Dportrait~\cite{trevithick2023real} reports that a minor misalignment between the ground truth and the estimated face poses will cause traditional pixel-based metrics to be unreliable. Recent work further indicates that deep perceptual image metrics are also sensitive to small misalignments~\cite{ghildyal2022shift}. Therefore, we additionally adopt the facial identity score $\text{ID}_{sim}$, $i.e.$ a metric to evaluate the preservation of the individual's identity~\cite{cao_pose-robust_2018,nitzan2022mystyle}. Nonetheless, we still include DISTS and LPIPS results in our inversion tasks to align with prior works~\cite{chan_efficient_2022, trevithick2023real}.
 
The identity score $\text{ID}_{sim}$ of an image is determined by the individual's reference set, which contains all the images available from that individual. Given an image, we extract the identity features and report the cosine similarity of the given image to the nearest image in the personal reference set. Formally, 
\begin{equation}
\text{ID}_{sim} \left ( w_i, c_i \right ) = max\left\{ \left \langle R ( G ( w_i, c_i, \cdot ) ) , R ( x_j ) \right \rangle \right\}_{j=1}^{N}
\label{eq:id_metric}
\end{equation}
where $R$ is a pretrained ArcFace~\cite{deng2019arcface} network for feature recognition, $\left<\cdot,\cdot\right>$ computes the cosine similarity between its argument as the ID scores~\cite{patashnik2021styleclip}, and $N$ is the number of 2D images in a reference set.
We report $\text{ID}_{sim}$ for personal identity preservation evaluation in all of our downstream applications, specifically in synthesis and interpolation tasks where there is no ground truth.
\noindent
\begin{wrapfigure}{r}{0.5\linewidth}
    \begin{minipage}{\linewidth}
        \vspace{-2em}
        \centering
        \includegraphics[width=\linewidth]{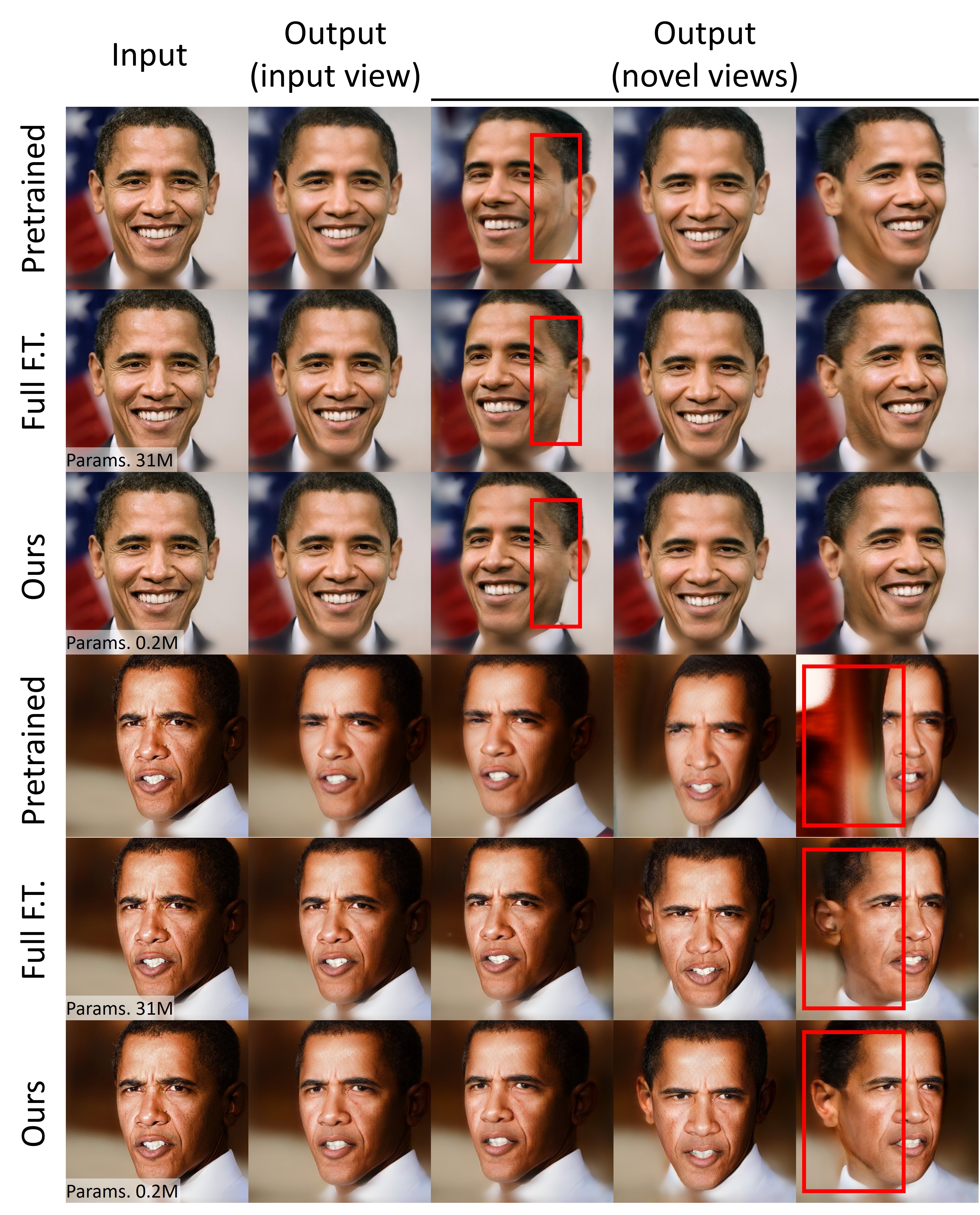} 
        \vspace{-2em}
        \caption{
        Qualitative evaluation for image inversion, $i.e.$ generating 3D-aware view synthesis from a single input image. Ours is My3DGen with LoRA rank $r = 1$ where the number of trainable parameters $= 0.2$M. Visual differences are highlighted with a red-box, zoom in to view finer details. 
        }
        \label{fig:inversion_multi_view}

        \centering
        \captionof{table}{Quantitative evaluation for image inversion. F.T. indicates fine-tuning, and Personal. Params. indicates the number of personalized parameters. DISTS and LPIPS compare the inverted image against the input image in the same pose, $\text{ID}_{sim}$ evaluates the preservation of identity across multiple poses. }
        \resizebox{\linewidth}{!}{%
        \begin{tabular}{cccccr}
        \toprule
    \multirow{2}{*}{\makecell[c]{F.T.}} & \multirow{2}{*}{\makecell[c]{LoRA}} & \multirow{2}{*}{\makecell[c]{DISTS $\downarrow$}} & \multirow{2}{*}{\makecell[c]{LPIPS $\downarrow$}} & \multirow{2}{*}{\makecell[c]{ID$_\text{sim} \uparrow$}} & \multirow{2}{*}{\makecell[l]{Personal. \\ Params. $\downarrow$}} \\
    & & & & & \\
        \cmidrule(lr){1-6} 
        - & - & 0.12 & 0.19 & 0.56 & - \\ 
        \checkmark & - & 0.08 & 0.12 & 0.60 & 31M\\ 
        \checkmark & \checkmark & 0.08 & 0.13 & 0.61 & 0.2M\\
        \bottomrule
        \end{tabular}%
        }

        \vspace{-0.5em}
        \vspace{-1.0em}
    \end{minipage}
    
\end{wrapfigure}

\vspace{-2.5em}
\subsection{Applications of My3DGen}
\vspace{-0.5em}
\label{exp:downstream}
\noindent
\textbf{Inversion.}
\label{exp:inversion}
Given a 2D RGB image of a face, image inversion here refers to performing 3D reconstruction of the face. We apply PTI~\cite{roich2022pivotal} for inversion to align with the original EG3D work~\cite{chan_efficient_2022}. We optimize a randomly initialized latent code for 600 iterations, followed by fine-tuning the model for an additional 350 iterations~\cite{trevithick2023real}. 
We calculate DISTS and LPIPS for single-view inversion, and report $\text{ID}_{sim}$ to evaluate multi-view reconstruction.
As shown in Fig.~\ref{fig:inversion_multi_view}, the pretrained EG3D can have artifacts such as i) a visible seam between the face and the rest of the head (1st row, 3rd column), ii) 3D shape distortion (4th row, 5th column), and iii) identity drift caused by changing pose, which can be observed in Fig.~\ref{fig:teaser} regarding novel view synthesis. Furthermore, the pretrained EG3D overly smooths skin textures. We recommend readers to zoom in to observe these finer details (2nd column of Fig.~\ref{fig:inversion_multi_view}). In contrast, ours decomposes facial features into global features and personalized features, producing a higher quality and identity-preserving inversion.
Compared to full fine-tuning, our approach yields almost the same quality for single-view inversion as measured by LPIPS, and produces similar multiview inversion results with a slightly higher $\text{ID}_{sim}$. 
We speculate that full fine-tuning can result in better single-view quality by inverting images with an overfitted background, but may also introduce artifacts by overfitting particular facial features. 
For background overfitting, we examine this problem in Fig.~\ref{fig:ablation_lora_rank} where we demonstrate that increasing LoRA rank enhances background fitting but provides minor improvements in terms of $\text{ID}_{sim}$. 
For facial features, as shown in Fig.~\ref{fig:inversion_multi_view} (5th row), full fine-tuning does not preserve the identity in the lower jaw and ear regions, which are obscured in the input image. Personalizing the model with LoRA can help reduce these artifacts, even including rendering `floaters' from NeRF~\cite{xie_s-nerf_2023}. We suggest the readers check the inversion video results in the supplementary for more examples.

\begin{wrapfigure}{r}{0.5\linewidth}
  \centering
  \vspace{-2.3em}
   \includegraphics[width=\linewidth]{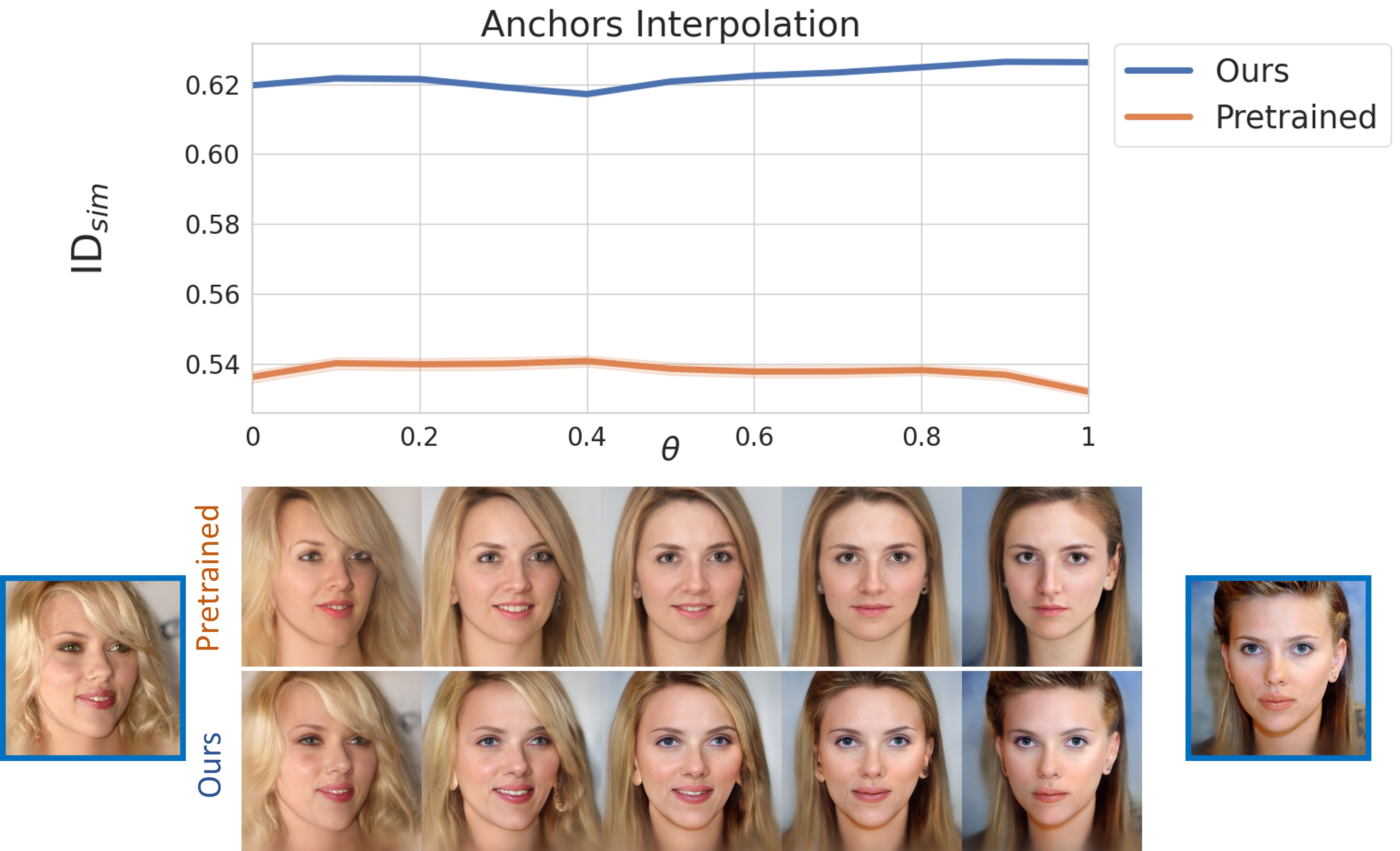}
   \vspace{-2em}
  \caption{
   Quantitative (top) and qualitative (bottom) evaluation for interpolation in latent space between two anchor images, highlighted in {\color{cyan}color}. We measure identity preservation using $\text{ID}_{sim}$, for which illustrative visual results at different interpolation steps $\theta$ are also provided. We compare the results {\color{orange}before} and {\color{blue}after} personalization.
   }
   \label{fig:interpolation}

    \begin{minipage}{\linewidth}
        \centering
        \includegraphics[width=\linewidth]{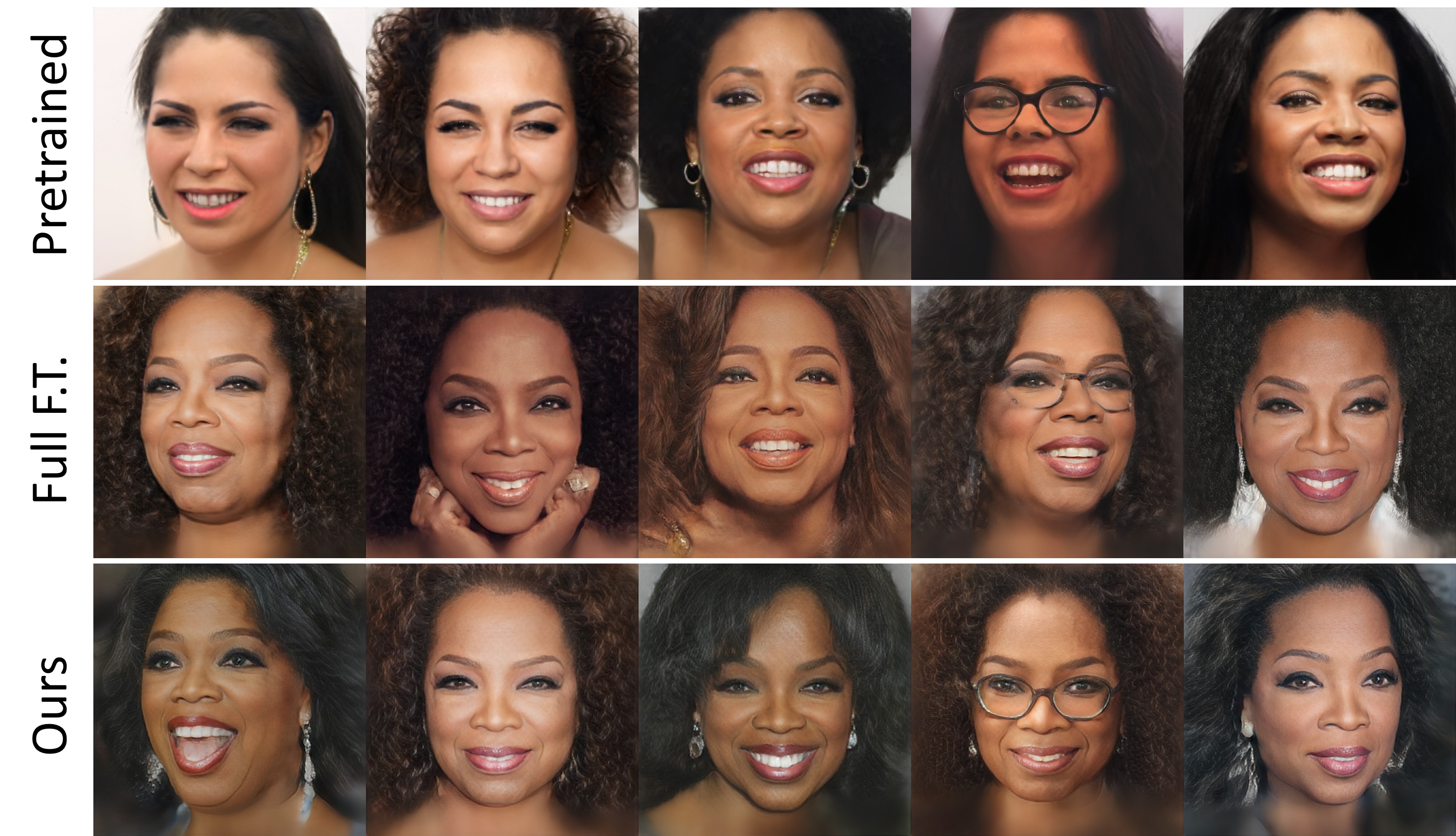}
        \vspace{-2em}
        \caption{Qualitative evaluation for synthesizing novel appearances of an individual.  Generated images of \textit{Oprah Winfrey} are provided for visual inspection.
        }
        \label{fig:synthesis}
    \vspace{-1em}
    
    \centering
    \captionof{table}{Quantitative evaluation for novel appearance synthesis. F.T. indicates fine-tuning and Personal. Params. indicates the number of personalized parameters.}
    \label{tab:synthesis}
    \resizebox{0.85\linewidth}{!}{%
    \begin{tabular}{ccccr}
        \toprule
        \multirow{2}{*}{\makecell[c]{F.T.}} & \multirow{2}{*}{\makecell[c]{LoRA}} & \multirow{2}{*}{\makecell[c]{ID$_\text{sim} \uparrow$}} & \multirow{2}{*}{\makecell[c]{$\text{Diversity}$ $\uparrow$}} & \multirow{2}{*}{\makecell[l]{Personal. \\ Params. $\downarrow$}} \\
        & & & & \\
        \cmidrule(lr){1-5}
        - & - & 0.53 & 0.19 & - \\ 
        \checkmark & - & 0.62 & 0.21 & 31M\\ 
        \checkmark & \checkmark & 0.62 & 0.21 & 0.2M\\ 
        \bottomrule
    \end{tabular}%
    }
    \vspace{-1em}
    \vspace{-2em}
    
    \end{minipage}
   
\end{wrapfigure}
\vspace{-0.25em}
\noindent
\textbf{Interpolation.}
\label{exp:interpolation}
The results obtained from the inversion tasks are overfitted to the test image via PTI, requiring separate optimization of the generator for each image. Our approach aims to provide a single generative model for all images of an individual without requiring further optimization. We support this claim by generating images from latent vectors produced through linear interpolation between two randomly selected training anchors~\cite{nitzan2022mystyle}.
We interpolate between the anchor pair at 10 equally spaced interpolation weights. At each interpolation step $\theta$, given the interpolated latent code, we randomly generate 20 novel views and compute the average ID$_{sim}$ scores of each view as described in Eq.~\ref{eq:id_metric}. We repeat this process for 10 randomly selected anchor pairs for each personalized model, averaging our results across all models and pairs. We compare the results before and after personalization in Fig.~\ref{fig:interpolation} and show that personalization maintains the identity when traversing between two anchor images (exemplified by~\textit{Scarlett Johansson}) while the pre-trained model distorts identity.
\noindent
\textbf{Synthesis.}
We conduct image synthesis to further examine the personalization capacity of our approach. Our goal is to produce new, distinctive images of an individual that have not been seen before. We sample a latent code from the convex hull following MyStyle~\cite{nitzan2022mystyle}. Then we feed the latent code into the generator together with a random pose. To this end, we randomly synthesize images from each generator and compare the results based on identity preservation.
We evaluate the image synthesis results and present them in Table~\ref{tab:synthesis}, accompanied by visual examples in Fig.~\ref{fig:synthesis}.
To assess the diversity of the synthesized images, we follow the protocol proposed by Ojha et al.~\cite{ojha2021few}. Specifically, for each celebrity, we produce 1,000 images and assign each of them to one of the $k$ training images, by choosing the one with the lowest LPIPS distance. We then compute the standard deviation of pairwise LPIPS distances within members of the same cluster and then average over the $k$ clusters.
Our model surpasses pretrained EG3D in all metrics and performs comparably to full fine-tuning with fewer trainable parameters.

\begin{wrapfigure}{r}{0.5\linewidth}
\vspace{-2em}
    \begin{minipage}{\linewidth}
        \centering
        \includegraphics[width=\linewidth]{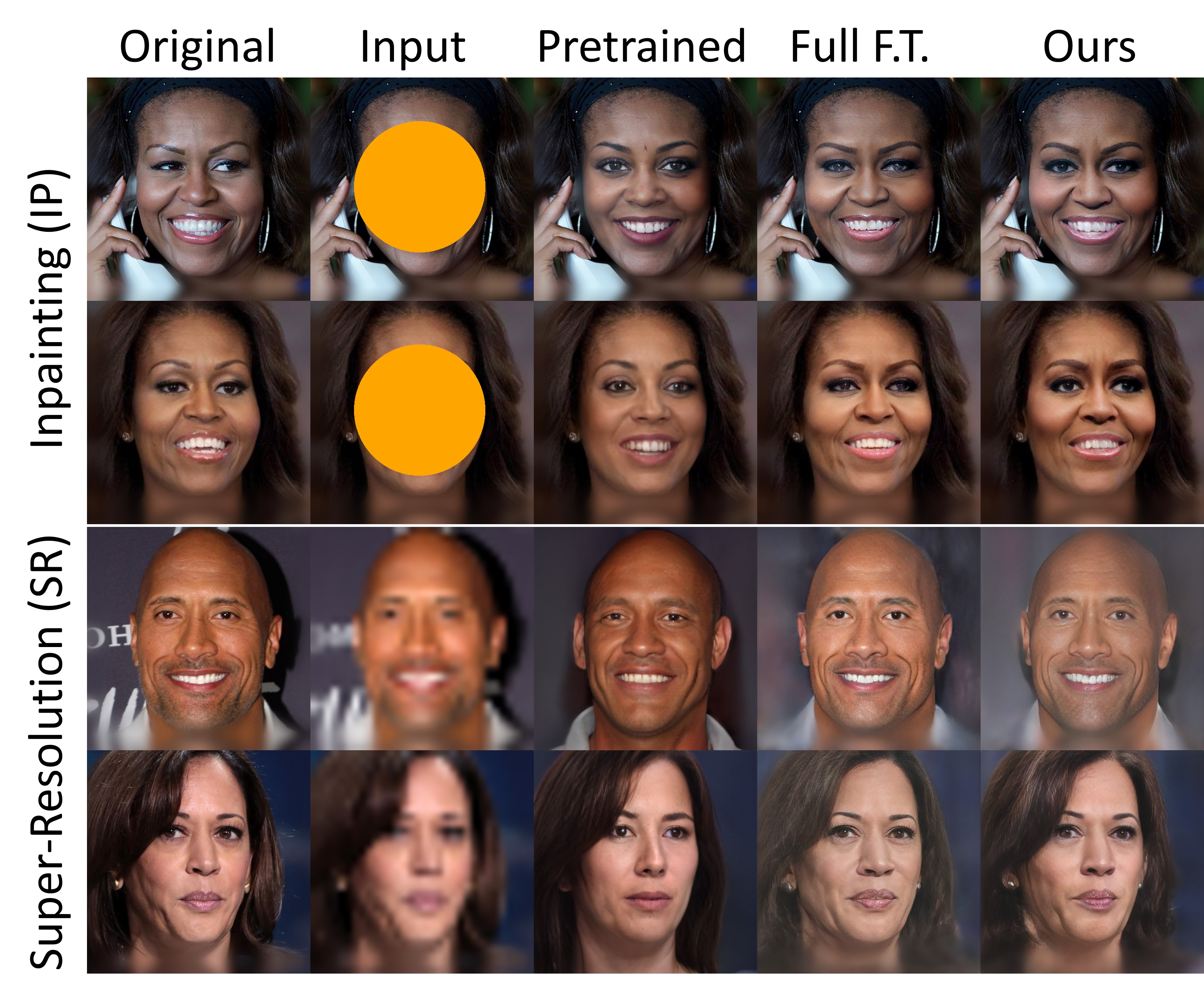}
        \vspace{-2em}
        \caption{Qualitative evaluation for image enhancement by inpainting (IP) and super-resolution (SR). The original images are degraded as input images and then fed into the model. We have included original images for the benefit of readers who may not be familiar with the faces of \textit{Michelle Obama, Dwayne Johnson, and Kamala Harris}. 
        }
        \label{fig:enhancement}
        \vspace{-1em}
        \centering
        \captionof{table}{Quantitative evaluation for image enhancement. F.T. indicates fine-tuning and Personal. Params. indicates the number of personalized parameters. $\text{User}~\%$ reflects the percentages of responses for each option in the image enhancement tasks.}
        \label{tab:enhancement}
        \resizebox{\linewidth}{!}{%
        \begin{tabular}{cccccr}
            \toprule
            
            \multirow{2}{*}{\makecell[c]{}} & \multirow{2}{*}{\makecell[c]{F.T.}} & \multirow{2}{*}{\makecell[c]{LoRA}} & \multirow{2}{*}{\makecell[c]{$\text{ID}_{\text{sim}}$ $\uparrow$}} & 
            \multirow{2}{*}{\makecell[c]{$\text{User}~\%$ $\uparrow$}} &
            \multirow{2}{*}{\makecell[l]{Personal. \\ Params. $\downarrow$}} \\
            & & & & & \\
        
            
            \cmidrule(lr){1-6} 
            
            \multirow{3}{*}{IP} & - & - &  0.62 & 12.7 & - \\ 
            &\checkmark & - & 0.72 & - & 31M\\ 
            &\checkmark & \checkmark & 0.72 & 81.8 & 0.2M\\ 
            \cmidrule(lr){1-6} 
            \multirow{3}{*}{SR} & - & - & 0.61 & 7.9 & - \\ 
            &\checkmark & - & 0.73 & - & 31M \\ 
            &\checkmark & \checkmark & 0.73 & 88.5 & 0.2M\\ 
            \bottomrule
        \end{tabular}%
        }
        \vspace{-0.5em}
        \vspace{-2em}
    \end{minipage}
\end{wrapfigure}
\vspace{-0.5em}
\noindent
\textbf{Image Enhancement.} 
\label{exp:image_enhancement}
We choose the tasks of image inpainting (IP) and super-resolution (SR) as representative examples of image enhancement.
To perform inpainting, we utilize a mask positioned at the center of the faces and feed the masked images into the generator. We subsequently post-process the resulting outputs by blending the generator's output within the masked area with the input in other areas. The background regions of the outputs are replaced through post-processing using a Lanczos-upsampled version of the input image. These regions have been segmented according to the methodology in~\cite{wadhwa2018synthetic}. For SR, we reduce the original image size from $512  \times512$ to $32\times32$ and supply the blurred images to the generator.
It is noteworthy that in previous studies on image enhancement~\cite{zhao_large_2021}, no quality evaluation is performed against a reference standard, $i.e.$ a ground truth image.
``Quality" here refers to a pixel-to-pixel comparison against the original ground truth. The reasoning for this is that although the restored facial details may differ from those of the original image, personalization still restores the key facial characteristics of the individual, resulting in valid restorations~\cite{nitzan2022mystyle, mystyle++}.
We adhere to the established convention and do not include quality evaluations either.
Therefore for completeness, in addition to using the ID$_{sim}$ metric, we further perform user studies that rely on subjective assessments of identity preservation in the experiments. We used Amazon Mechanical Turk to gather 330 responses from 17 users. In the study, users were shown the original image, 
along with two results: one from the pretrained model and the other from the personalized model, in a randomized order. Users were instructed to select the result that most closely resembled the person in the picture and maintained high fidelity to the original input. In cases where both results were similar, participants could select `No Preference'. Results are reported as a percentage of each selected option. Images utilized in the user study were selected from a random subset of those used for quantitative evaluation.
We present the enhancement results in Fig.~\ref{fig:enhancement} and Table~\ref{tab:enhancement}, showing visual examples of inpainting (IP) and super-resolution (SR) tasks followed by both quantitative and qualitative analysis. The User~\% for `No Preference' is $\text{5.5}\%$ for inpainting and $\text{3.6}\%$ for super-resolution.
Our results demonstrate that My3DGen successfully integrates personal features into the pretrained EG3D with both higher quantitative score (ID$_{sim}$), and qualitative preference (User~\%).

\begin{wrapfigure}{r}{0.5\linewidth}
    \vspace{-2.5em}
    \begin{minipage}{\linewidth}
        \centering
        \includegraphics[width=\linewidth]{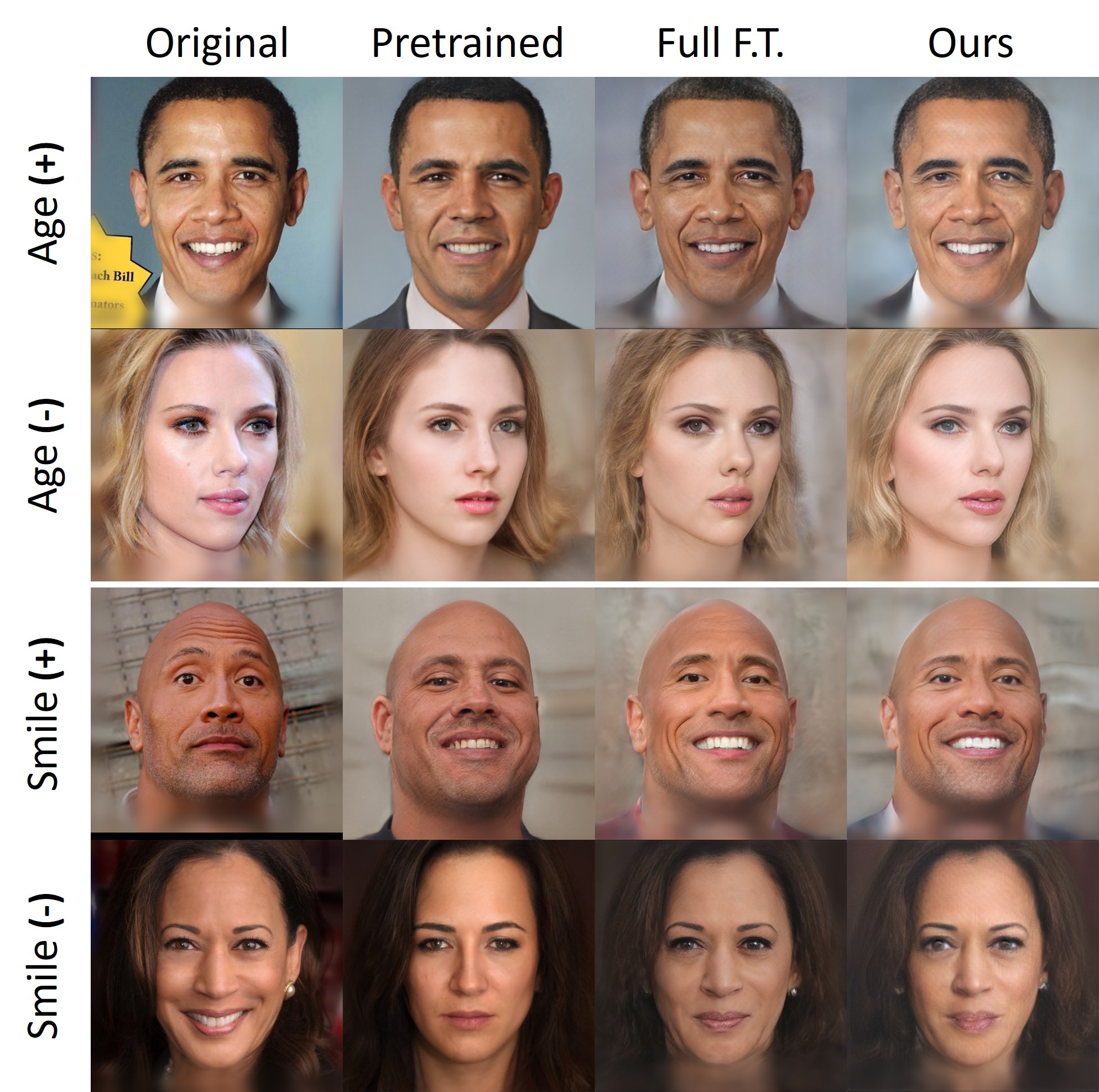}
    \vspace{-2em}
    \caption{Qualitative evaluation for semantic editing of `smile' and `age'.
    We show \textit{Barack Obama, Scarlett Johansson, Dwayne Johnson, and Kamala Harris} top-to-bottom.
    }
    \label{fig:semantic_editing}
    
    
    \centering
    \vspace{-1em}
    \captionof{table}{Quantitative evaluation for semantic editing.
    F.T. indicates fine-tuning and Personal. Params. indicates the number of personalized parameters. $\text{User}~\%$ reflects the percentages of responses for each option.}
    \label{tab:semantic_editing}
    \resizebox{0.85\linewidth}{!}{%
    \begin{tabular}{ccccr}
        \toprule
        \multirow{2}{*}{\makecell[c]{F.T.}} & 
        \multirow{2}{*}{\makecell[c]{LoRA}} & 
        \multirow{2}{*}{\makecell[c]{$\text{ID}_{\text{sim}}$ $\uparrow$}} &
        \multirow{2}{*}{\makecell[c]{$\text{User}~\%$ $\uparrow$}} &
        \multirow{2}{*}{\makecell[l]{\makecell[l]{Personal.} \\ \makecell[l]{Params. $\downarrow$}}} \\
        & & & & \\
        \cmidrule(lr){1-5}
        - & - &  0.60 & 4.8 & -\\ 
        \checkmark & - & 0.76 & - & 31M \\ 
        \checkmark & \checkmark & 0.76 & 95.2 & 0.2M \\ 
        \bottomrule
    \end{tabular}%
    }
    \vspace{-0.5em}
    \vspace{-2.5em}
    \end{minipage}
\end{wrapfigure}

\vspace{-0.5em}
\noindent
\textbf{Semantic Editing.}
\label{exp:semantic_editing}
Besides enhancing identity preservation in reconstruction tasks, personalization also enables tailored editing of facial attributes. For example, while a generalized pre-trained model may learn how an average human smiles or ages with time, it is not able to capture the specific smile or aging process of an individual well.  Our hypothesis is that a personalized generative prior can enable us to more accurately depict how specific attributes of an individual change. To validate this hypothesis, we perform semantic editing of `smile' and `age' features using 3D generative priors, both pretrained and personalized. We start by identifying suitable editing directions for each model using the InterFaceGAN framework~\cite{shen2020interpreting}, followed by reconstructing the input image and then conducting personalized semantic editing in the $\alpha$-space, as described in MyStyle~\cite{nitzan2022mystyle}. 
Our findings are illustrated in Table~\ref{tab:semantic_editing}, where we also employ the user study as quantitative metrics for evaluation. Our study design for the user survey aligns with that of the image enhancement evaluations, including 165 responses from a total of 17 participants.
Quantitatively, our model yields a nearly identical $\text{ID}_{sim}$ to the fully fine-tuned EG3D model and substantially higher $\text{ID}_{sim}$ than the pretrained model.
Qualitatively, we note that our model can sometimes outperform the fully fine-tuned model in retaining image styles, such as lighting, background color, and hairstyle post-editing. A detailed examination of Fig.~\ref{fig:semantic_editing} demonstrates improved preservation of skin tone and background color in the first row, enhanced reproduction of hair color, style, and background color in the second row, and better handling of lighting and skin tone in the last row, showing LoRA may mitigate overfitting in semantic editing.

\vspace{-1em}
\subsection{Analysis of Personalization with LoRA}
\label{exp:lora_weight_change}
In Sec.~\ref{exp:semantic_editing}, we show that personalizing 3D-GAN with LoRA can help avoid overfitting and has the potential to improve downstream results such as inversion and semantic editing compared to naive full fine-tuning.
In this section, we start by demonstrating that tuning the pre-trained model with LoRA rank $r=1$ suffices for personalization. We then study the importance of feature blocks with different resolutions for personalization with rank 1.

\begin{wrapfigure}{r}{0.5\linewidth}
\vspace{-2em}
    \begin{minipage}{\linewidth}
    \centering
    \captionof{table}{Quantitative analysis of the rank of LoRA for inversion and interpolation tasks. `-' indicates full fine-tuning. We report the average $\text{ID}_{sim}$ across latent paths from the interpolation task.}
    \label{tab:ablation_lora_rank}
    \resizebox{\linewidth}{!}{%
    \begin{tabular}{lccccr}
    \toprule 
    & \multicolumn{2}{c}{Inversion} & & \multicolumn{1}{l}{Interpolation} & \multicolumn{1}{l}{Personal.} \\
    \cmidrule(lr){2-3} \cmidrule(lr){5-5}
    LoRA &
    LPIPS $\downarrow$ & ID$_\text{sim} \uparrow$ & &
    ID$_\text{sim} \uparrow$ & Params. $\downarrow$\\
    \cmidrule(lr){1-6}
    \multicolumn{1}{c}{\centering-} & 0.12 & 0.60 & & 0.62 & 31M \\
    rank=1 & 0.13 & 0.60 & & 0.62 & 0.2M \\
    rank=4 & 0.11 & 0.60 & & 0.63 & 0.9M \\
    rank=16 & 0.09 & 0.60 & & 0.63 & 3.5M \\
    \bottomrule
    \vspace{0.5em}
    \end{tabular}%
    }
    \centering
    \includegraphics[width=\linewidth]{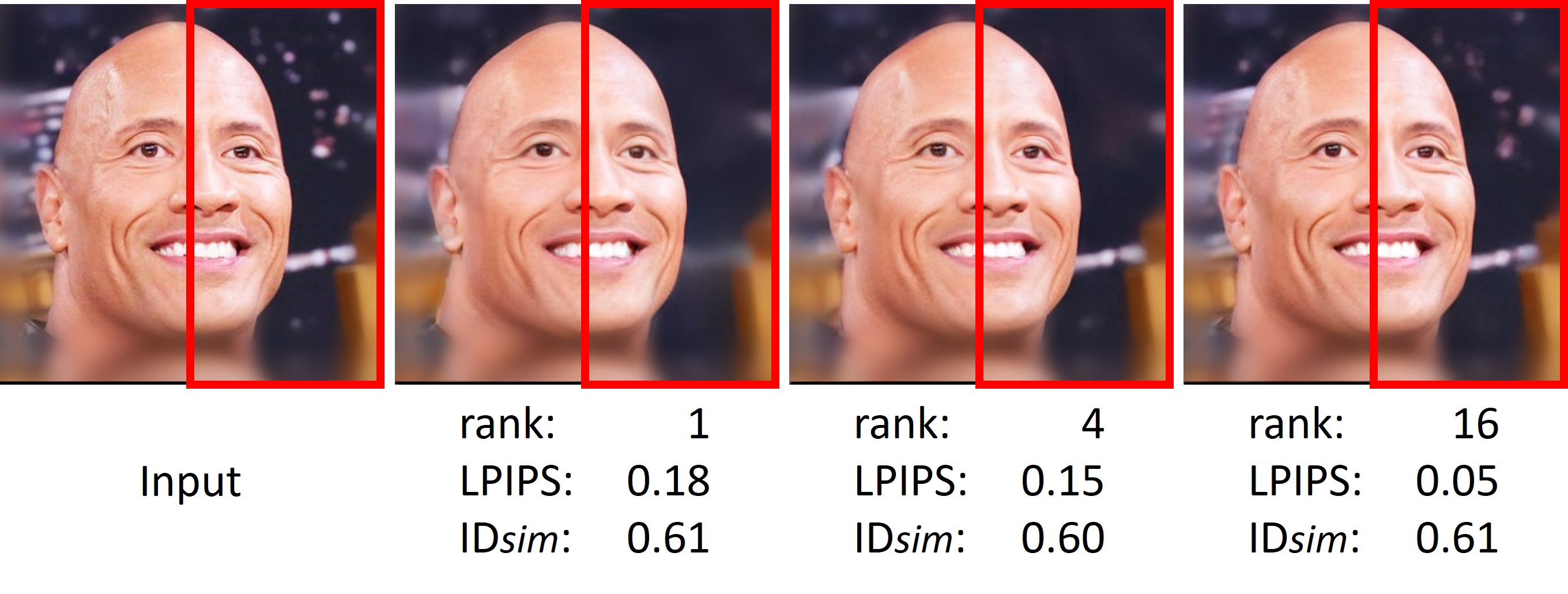}
    \vspace{-2em}
    \captionof{figure}{Visual examples of inversion results for different LoRA ranks. The backgrounds are highlighted to aid inspection along with LPIPS and $\text{ID}_{sim}$ scores.}
    \label{fig:ablation_lora_rank}
    \end{minipage}

\end{wrapfigure}

\noindent
\textbf{Effect of Rank in LoRA.}
To investigate the impact of rank selection in LoRA and find the optimal rank, we repeat the same personalization method for each celebrity, with LoRA ranks of 1, 4, and 16, tuning the generator accordingly. 
We report the model's performance on inversion and interpolation tasks in Table~\ref{tab:ablation_lora_rank}.
For inversion tasks, increasing the rank from 1 to 16 doesn't help improve identity preservation measured by $\text{ID}_{sim}$,
but it does help to fit the background and thus leads to a better LPIPS score, for which visual results are shown in Fig.~\ref{fig:ablation_lora_rank}.

This clarifies the contrast between our method and full fine-tuning regarding single-view inversion results outlined in the inversion task in Sec.~\ref{exp:inversion}.
However, matching backgrounds is not the primary focus of personalization, so the slight performance tradeoff is acceptable in this case.
Additionally, there is only a slight improvement in $\text{ID}_{sim}$ for interpolation tasks when the rank increases. Thus, we choose to use rank $=1$ in most of our experiments.

\noindent
\textbf{Personalization of Feature Blocks.}
We show that feature blocks with different resolutions have different levels of importance during personalization. 
Determining the importance of weights in a network is an open question~\cite{blalock_what_2020}. Previous work proposes including gradient information to evaluate the importance of LoRA modules during the tuning~\cite{zhang_adaptive_2023}. Here, after tuning with LoRA, we follow the common approach in model pruning where the change in parameter magnitude is used as a criterion to evaluate layer importance~\cite{lee_layer-adaptive_2021, elesedy_lottery_2021}. That is, $|\Delta W| /  |W_{0}| \times 100\%$, where $\Delta W$ is the personalized LoRA weights and $W_{0}$ is the pre-trained weights. In EG3D, the first 7 resolution blocks of StyleGAN2 are used as the backbone to generate tri-plane representations of $256\times256$ resolution. For each resolution block, we calculate the relative LoRA weight change percentage compared to the pre-trained weight.

\begin{wrapfigure}{r}{0.5\linewidth}
\vspace{-1em}
    \centering
    \includegraphics[width=\linewidth]{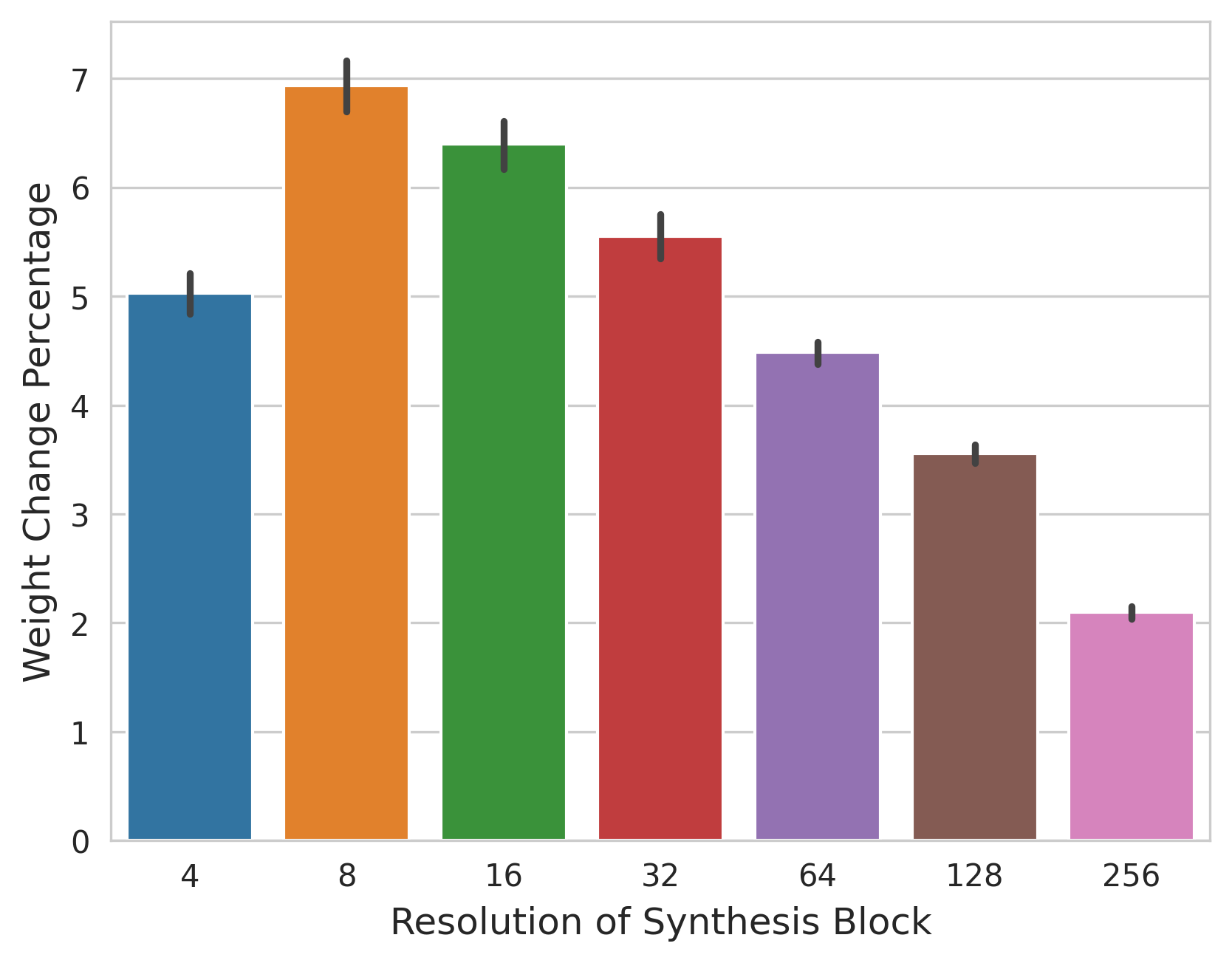}
    \vspace{-1.5em}
    \caption{LoRA weight change compared to the pre-trained weights after personalization, averaged across celebrities.}
    \label{fig:lora_weight_change}
    \vspace{-3em}
\end{wrapfigure} 

We report the mean and variance of the changes across different individuals in Fig.~\ref{fig:lora_weight_change}. The results show that feature blocks of resolution $8\times8$, $16\times16$, and $32\times32$ require more weight changes than other resolution blocks. 
This indicates that personal facial features are learned through the `coarse' and `middle' layers during personalization while little changes are required for `fine' layers. This finding aligns with the results in StyleGAN2~\cite{karras2019style} that `fine' layers affect only micro-features and finer details in the image such as color schemes and hairstyles.



\subsection{Ablation study}
\vspace{-0.5em}
\label{exp:ablation}

\noindent
\textbf{Effect of Dataset Size.}
\noindent
We investigate the effect of training dataset size on the performance of the personalized generator. We sample various random subsets of images for each celebrity, with sizes of 10, 50, and 100, and tune the generator accordingly. Next, we assess the generator's performance using the interpolation tasks outlined in Section~\ref{exp:interpolation},
which evaluates the generalization capacity of the personal convex hull by traversing through the latent subspace.
We report the average $\text{ID}_{sim}$ across latent paths.

\begin{table}[]
    \vspace{-2em}
    \centering
    \caption{The effect of training dataset size $\mathcal{D}_p$ on personalization. We evaluate personalization through the interpolation task in Sec.~\ref{exp:interpolation} and report the average $\text{ID}_{sim}$.}
        \begin{tabular}{ccccc}
            \toprule
            \multirowcell{2}[-0.4ex]{Dataset $\mathcal{D}_p$ \\ Size Ablations}
            & Metric & $\mathcal{D}_p = 10$ & $\mathcal{D}_p = 50$ & $\mathcal{D}_p = 100$ \\
            \cmidrule(lr){2-5}
            & 
            ID$_\text{sim} \uparrow$
            &
            0.618
            &
            0.628
            &
            0.629
            \\
            \bottomrule
        \end{tabular}

    \label{tab:ablation_dataset_size}
\end{table}

As shown in Table~\ref{tab:ablation_dataset_size}, the performance improves significantly from 10 to 50 images, but there is no significant improvement from 50 to 100 images. 
While further experiments are required, we speculate that adding more images may not contribute to dataset diversity and might even hurt the results~\cite{nitzan2022mystyle}. 
Therefore, we use 50 images for our personalization method in most of our experiments, unless otherwise stated.

\noindent
\textbf{AFHQv2 Cats.}
In addition to human faces, we also extend our personalization method to cat faces. Leveraging a photo album consisting of 22 in-the-wild images of one individual cat, we detect poses following~\cite{brad_kairesscat_hipsterizer_2024} and apply the same procedure used for human faces to personalize the pretrained EG3D-AFHQ model, which was pretrained on a dataset of 15000 animal images, including 5000 cat images across different identities and breeds.

\begin{wrapfigure}{r}{0.5\linewidth}
\vspace{-2em}
\centering
 \includegraphics[width=\linewidth]{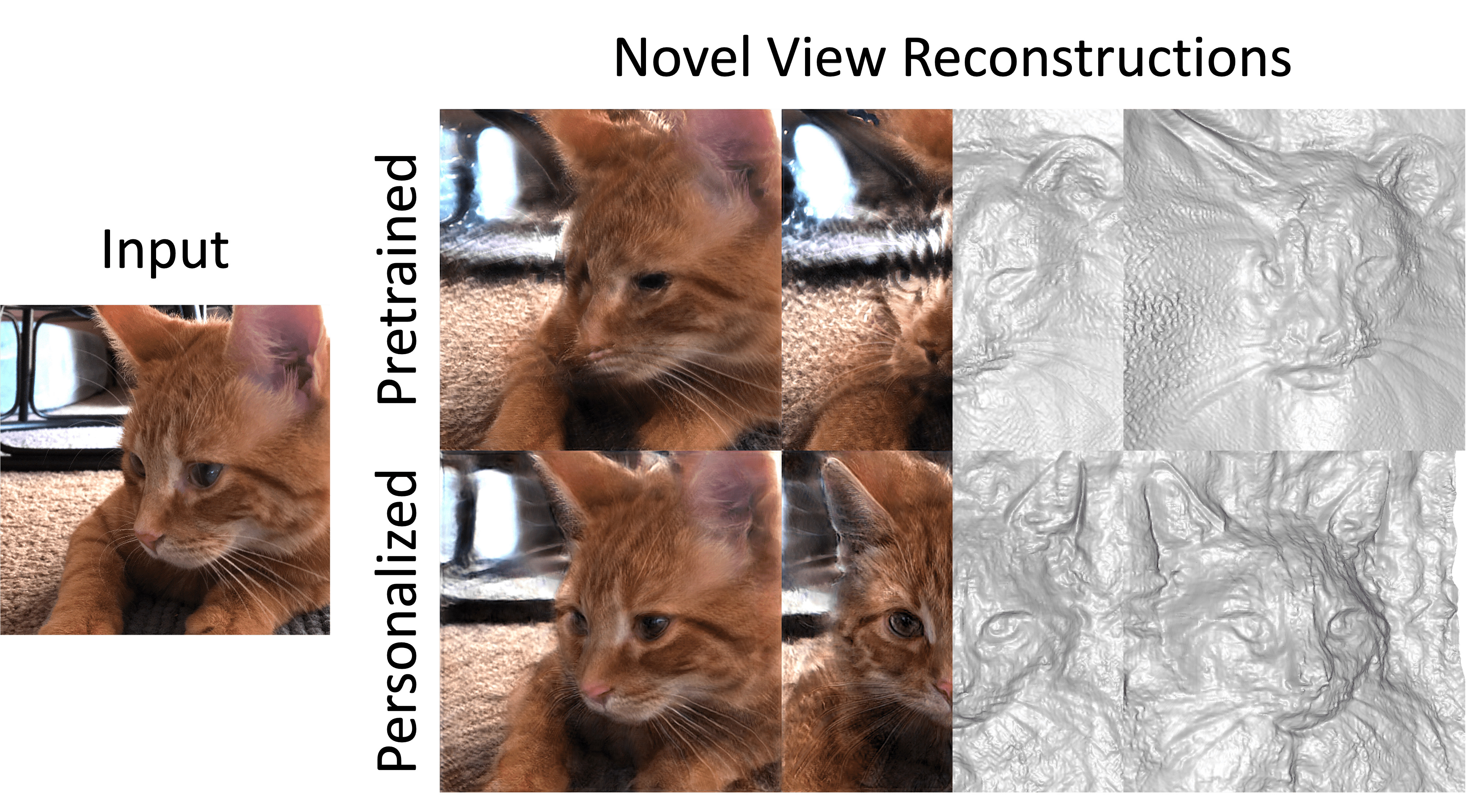}
\caption{Comparison between a pre-trained model and a personalized model for inverting an in-the-wild cat photo.}
\vspace{-2em}
\label{fig:cat}
\end{wrapfigure}

The results, showcased in Fig.~\ref{fig:cat}, demonstrate that our personalization technique significantly enhances the quality of the pretrained triplane representations for cat faces. 
This successful extension of our approach demonstrates the versatility and effectiveness of our method across different domains, paving the way for personalized 3D generative modeling of full human bodies, other animals, or objects.

\vspace{-0.5em}
\section{Discussion}
\noindent
\textbf{Limitations and Future Work.}
Our model can accurately capture facial features but encounters difficulty when objects heavily obscure the face (e.g. hats, phones, etc). Further works may leverage the power of EG3D-based encoder~\cite{yuan_make_2023, bhattarai2024triplanenet} for a better inversion performance.
Ours also struggles with heavily cropped faces, where the whole face is not fully captured in the original image and thus image boundaries are filled with reflection-padded values to align the face during preprocessing. One may mask out these invalid values during inversion. This will be similar to the in-painting task discussed in Section~\ref{exp:image_enhancement}.


\noindent
\textbf{Ethical Considerations.}
This study holds the potential to generate manipulated images of actual individuals, posing a substantial societal threat. Future research for detecting fake composites is needed.

\noindent
\textbf{Conclusion.}
We propose a parameter-efficient framework to personalize a large pretrained 3D generative model. Ours incorporates an individual's personal facial features into the pretrained model for downstream applications while preserving unique identity features. This will enable scalable personalization of 3D generative models in the real world. 
\clearpage  

%
%
\bibliographystyle{splncs04}
\bibliography{references_luchao, references_luchao_fixed, references_extra}

\clearpage
\setcounter{page}{1}
\appendix
Except for this supplemental PDF, we wrap other visual supplemental materials (images, videos, etc.) into an HTML file that can be viewed in \textit{result.html}. We highly recommend readers to refer to the accompanying videos for a comprehensive examination of the visual outcomes.
\vspace{-1em}
\section{Overview of Appendices}
\begin{wrapfigure}{r}{0.5\linewidth}
    \centering
    \vspace{-2em}
    \includegraphics[width=\linewidth]{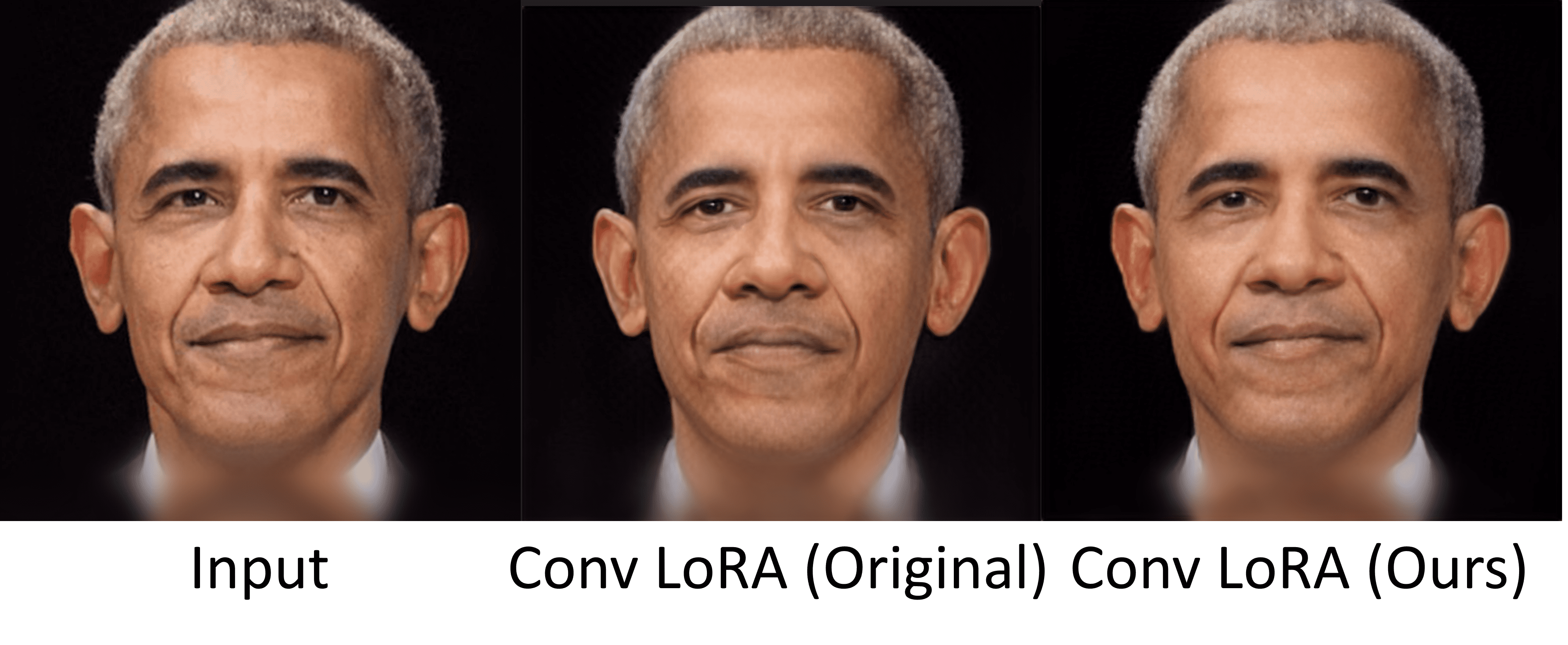}
    \vspace{-2em}
    \caption{
    Following the same personalization pipeline,
    we compare reconstructed results using the original LoRA code (middle) and our own LoRA implementation (right). 
    The original algorithm introduces idiosyncratic artifacts, such as diagonal stripe patterns.
    It is recommended to zoom in for finer details, especially around the cheeks and forehead.
    }
    \label{sup:lora_implement_diff}
    \vspace{-1em}
\end{wrapfigure}

\vspace{-1em}
Our appendices contain the following additional details:
\begin{itemize}[noitemsep,topsep=0pt]
    \item Sec.~\ref{sup:conv_lora} describes the details of our convolutional LoRA decomposition, where we show the difference between the original implementation and ours. We compare the results visually in Fig.~\ref{sup:lora_implement_diff}, where facial texture is accompanied by checkerboard artifacts using the original LoRA paper's code implementation~\cite{hu_lora_2021}.
    \item Sec.~\ref{sup:dataset_size} provides details on the celebrity dataset and additional information regarding our ablation studies on the effect of dataset size in Sec.~\ref{exp:ablation}.
    \item Sec.~\ref{sup:no_pti_inversion} shows image inversion results without PTI~\cite{roich2022pivotal} in Fig.~\ref{sup:no_pti_inversion_fig}. We re-project latent code into personal convex hull following Mystyle~\cite{nitzan2022mystyle} for inversion where model weights remain unchanged.
    \item Sec.~\ref{sup:interpolation_curve} further discusses the interpolation results shown in Sec.~\ref{exp:interpolation}, particularly why the interpolation curve is different from that in Mystyle~\cite{nitzan2022mystyle}.
    \item Sec.~\ref{sup:training_time} describes the detailed hardware configurations and training time for our experiments.
    \item In Sec.~\ref{sup:failure_sec}, we display failure cases for our experiments in Fig.~\ref{sup:failures}.
\end{itemize}





\section{LoRA for Convolutional Layer}
\label{sup:conv_lora}
\vspace{-1em}
LoRA~\cite{hu_lora_2021} is originally defined for matrix multiplication for fully-connected layers. However, convolution operation with \(C_1\) output channels, \(C_2\) input channels, and kernel size of \(k \times k \) is often implemented as matrix multiplication with a matrix \(W\) under ``im2col''~\cite{chellapilla2006high} transform on the image \(X\). The matrix \(W\) has dimension \(W \in \mathbb{R}^{C_1 \times C_2 k k}\).
\setlength{\abovedisplayskip}{6pt}
\setlength{\belowdisplayskip}{6pt}
\begin{equation}
    \operatorname{Conv}_\theta (X) = W \operatorname{im2col}(X)
\end{equation}

Therefore, we can decompose matrix \(M\) for convolution layers similar to LoRA. With a rank \(r\) LoRA decomposition, let \(B \in \mathbb{R}^{C_1 \times r}\) and \(A \in \mathbb{R}^{r \times C_2 k k}\), we have the following equation.
\setlength{\abovedisplayskip}{3pt}
\setlength{\belowdisplayskip}{3pt}
\begin{equation}
    W = B A
\end{equation}

We found official LoRA~\cite{hu_lora_2021} implementation performs the following decomposition. Matrix \(W\) is assumed to have dimension \(W \in \mathbb{R}^{C_1 k \times C_2 k }\), while \(B\), \(A\) have dimension \(B \in \mathbb{R}^{C_1 k \times r}\), \(A \in \mathbb{R}^{r \times C_2 k}\). 

\noindent
\\
Ours differs from the original LoRA implementation in two ways:
\begin{itemize}
    \item LoRA showed weight matrix \(W \in \mathbb{R}^{C_1 \times C_2 k k}\) that maps from input space \(C_2 k k\) to output space \(C_1\) of a layer can have low rank structure. It is unclear if matrix \(W \in \mathbb{R}^{C_1 k \times C_2 k }\) has low rank structure.
    \item We are surprised to find that the official implementation of LoRA directly interprets the memory content of the matrix \(W \in \mathbb{R}^{C_1 k \times C_2 k}\) as \(W \in \mathbb{R}^{C_1 \times C_2 k k}\) and perform convolution operation. We suspect this is a bug. Even though the matrix \(W\) is trainable, we suspect that such implementation has important consequences on performance, as now we are equivalently trying to find a low rank decomposition for a matrix that has no clear meaning, and might not have a low rank structure.
\end{itemize}

\noindent
We compare ours with the official implementation of LoRA in Fig.~\ref{sup:lora_implement_diff}, where the official implementation introduces checkerboard artifacts while our implementation is better at keeping the original image content.

\begin{wraptable}{R}{0.5\linewidth}
\vspace{-2em}
\centering
\captionof{table}{The sizes of the reference and test sets of our dataset.}
\resizebox{\linewidth}{!}{%
\begin{tabular}{lcc}
\hline \makecell[c]{Celebrity}
 & Reference set size & Test set size \\
\hline
Barack Obama & 192 & 13 \\
Dwayne Johnson & 97 & 12 \\
Joe Biden & 200 & 13 \\
Kamala Harris & 110 & 7 \\
Michelle Obama & 279 & 9 \\
Oprah Winfrey & 135 & 9 \\
Scarlett Johansson & 260 & 13 \\
Taylor Swift & 158 & 9 \\
Xi Jinping & 92 & 15 \\
\hline \hline
\end{tabular}%
}

\label{sup:dataset_table}
\vspace{-1em}
\end{wraptable}

\section{Dataset Size}
\label{sup:dataset_size}
Using the same dataset in Mystyle~\cite{nitzan2022mystyle}, we further process the images following the preprocessing pipeline in EG3D~\cite{chan_efficient_2022}.
We show the number of images in the reference and test sets in Table~\ref{sup:dataset_table}. In Sec.~\ref{exp:ablation}, we conduct ablation studies to investigate the effect of the size of the training set. When tuning on 100 images, if the reference  set size is below 100, we use all the images in the reference set as the training set, such as 97 for \textit{ Dwayne Johnson} and 92 for \textit{Xi Jinping}. Unless otherwise specified, we tune on 50 images for the majority of our personalization experiments.


\section{Image Inversion without PTI}
\label{sup:no_pti_inversion}
\begin{wrapfigure}{r}{0.5\linewidth}
\vspace{-2em}
\centering
\includegraphics[width=\linewidth]{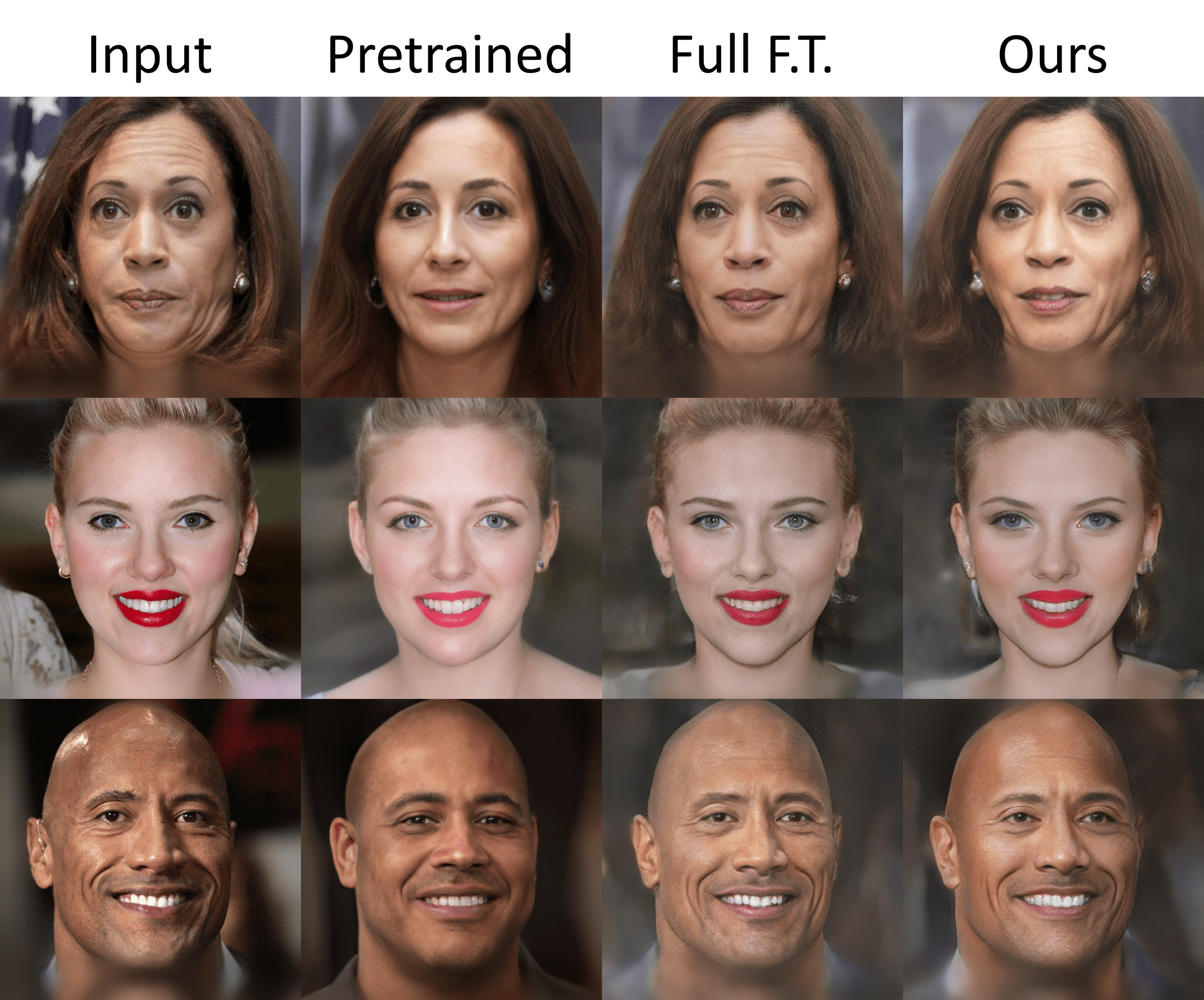}
\vspace{-1em}
\caption{
Image inversion results without optimizing network weights. F.T. indicates fine-tuning and ours is My3DGen.
}
\label{sup:no_pti_inversion_fig}

\includegraphics[width=\linewidth]{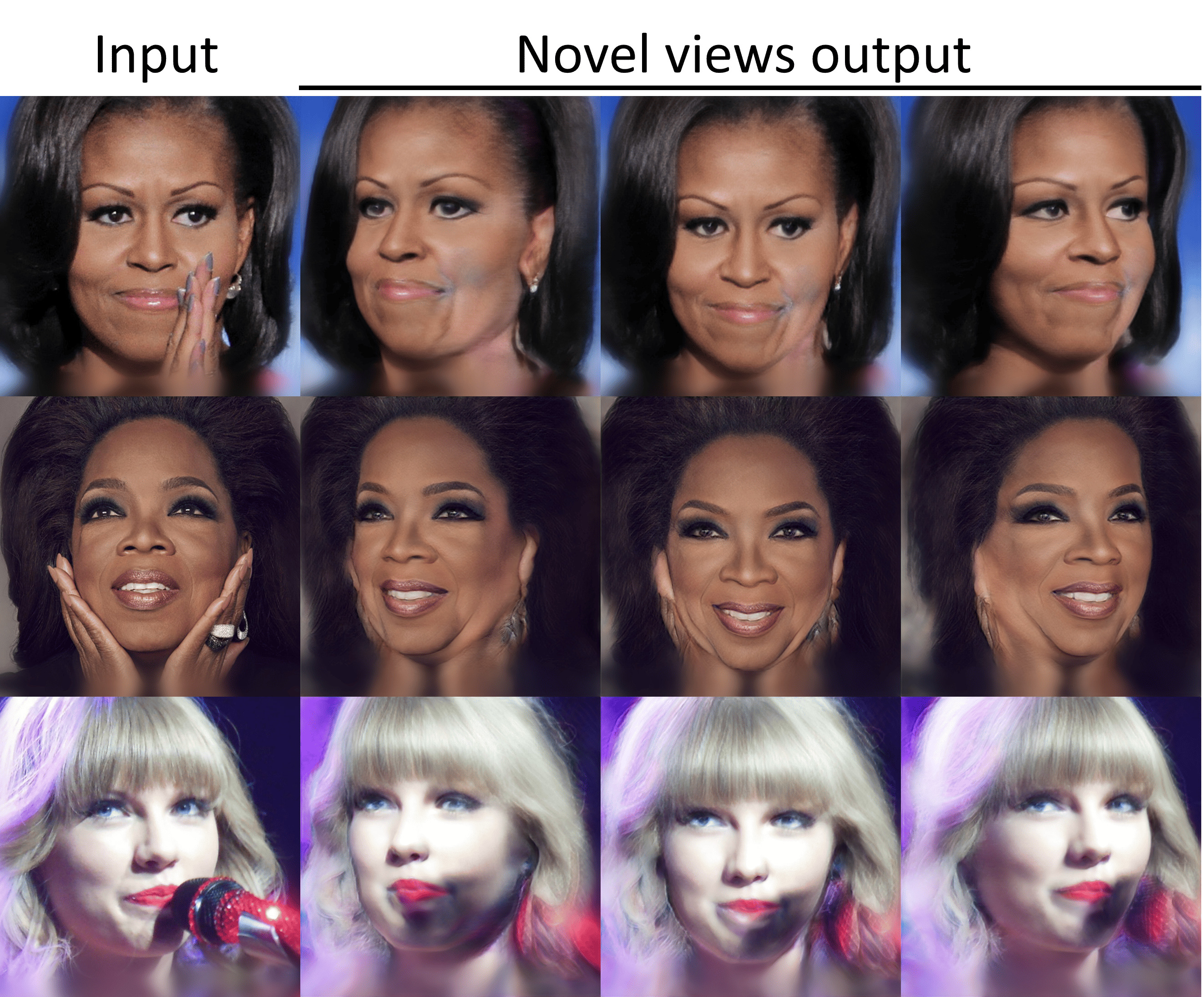}
\vspace{-1em}
\caption{
Cases where the inversion method fails to reconstruct objects.
}
\vspace{-1em}
\label{sup:failures}
\end{wrapfigure}

In Sec.~\ref{exp:inversion}, we perform image inversion tasks using PTI~\cite{roich2022pivotal} to align with previous works~\cite{chan_efficient_2022, trevithick2023real}, where PTI requires changing the model weights for the best inversion quality. Nevertheless, we provide inversion results following Mystyle~\cite{nitzan2022mystyle} where the model weights remain unchanged and only the latent code is re-projected into the convex hull. As shown in Fig.~\ref{sup:no_pti_inversion_fig}, although personalization helps maintain identity in the inversion tasks, it still lacks facial details for both full fine-tuning and ours, compared to PTI. Further works may design an encoder for EG3D inversion similar to TriPlaneNet~\cite{bhattarai2024triplanenet}.

\section{ID$_{sim}$ Curve Shape in Interpolation Tasks}
\label{sup:interpolation_curve}
\vspace{-1em}
Interestingly, unlike the previous findings of Mystyle~\cite{nitzan2022mystyle}, there is no significant difference in $\text{ID}_{sim}$ scores between the interpolated latent codes and the anchors.
The interpolation ID$_{sim}$ curve in Mystyle follows a reserved U-shape, while our curve is flatter, as shown in Fig.~\ref{fig:interpolation}.
It is hypothesized that this lack of difference may be due to both our $\text{ID}_{sim}$ metric design and EG3D's 3D advantage, where the extreme properties of anchors, such as pose, have a smaller impact on $\text{ID}_{sim}$ compared to 2D-GANs. 

\vspace{-0.5em}
\section{Training Time}
\label{sup:training_time}
\vspace{-0.5em}
We perform our personalization experiments on 4 NVIDIA RTX A6000 GPUs. Our total training time is 5 hours with LoRA, compared to 6 hours without LoRA.


\section{Failure Cases}
\label{sup:failure_sec}
\vspace{-0.5em}
My3DGen struggles to reconstruct objects that obscure the face, such as hands and phones, even with PTI. The cause for this is a deficiency in corresponding images of objects in the pre-training facial dataset, FFHQ. Further works may design an EG3D-specific encoder that can encode objects into the latent space similar to Live3DPortrait~\cite{trevithick2023real}.



\end{document}